\documentclass[letterpaper]{article}
\usepackage{aaai25} % DO NOT CHANGE THIS
\usepackage{times} % DO NOT CHANGE THIS
\usepackage{helvet} % DO NOT CHANGE THIS
\usepackage{courier} % DO NOT CHANGE THIS
\usepackage[hyphens]{url} % DO NOT CHANGE THIS
\usepackage{graphicx} % DO NOT CHANGE THIS
\usepackage{graphicx} % DO NOT CHANGE THIS
\usepackage{natbib} % DO NOT CHANGE THIS
\usepackage{caption} % DO NOT CHANGE THIS
\usepackage{aaai25}  % arXiv sub

\usepackage{algorithm}
\usepackage{algorithmic}
\usepackage{newfloat}
\usepackage{listings}
\DeclareCaptionStyle{ruled}{labelfont=normalfont,labelsep=colon,strut=off} % DO NOT CHANGE THIS
\floatstyle{ruled}
\newfloat{listing}{tb}{lst}{}
\floatname{listing}{Listing}

\usepackage{amssymb}
\usepackage{amsmath}
\usepackage{float} % 使用float包来控制表格位置
\usepackage[hyphens]{url}
\usepackage{graphicx}
\usepackage{array}
\usepackage{multirow}

\begin{document}
% The file aaai.sty is the style file for AAAI Press 
% proceedings, working notes, and technical reports.
%
\title{From Words to Worth: Newborn Article Impact Prediction with LLM}
\author {
    % Authors
    Penghai Zhao\textsuperscript{\rm 1},
    Qinghua Xing\textsuperscript{\rm 1},
    Kairan Dou\textsuperscript{\rm 1},
    Jinyu Tian\textsuperscript{\rm 1},
    Ying Tai\textsuperscript{\rm 3},
    Jian Yang\textsuperscript{\rm 1}, \\
    Ming-Ming Cheng\textsuperscript{\rm 12},
    Xiang Li\textsuperscript{\rm *12}
}
\affiliations {
    % Affiliations
    \textsuperscript{\rm 1}VCIP, CS, Nankai University \ \ 
    \textsuperscript{\rm 2}NKIARI, Shenzhen Futian \ \
    \textsuperscript{\rm 3}PCALab, Nanjing University \\
   zhaopenghai@mail.nankai.edu.cn, xiang.li.implus@nankai.edu.cn
}

\maketitle

\begin{abstract}
\begin{quote}
As the academic landscape expands, the challenge of efficiently identifying impactful newly published articles grows increasingly vital. This paper introduces a promising approach, leveraging the capabilities of LLMs to predict the future impact of newborn articles solely based on titles and abstracts. Moving beyond traditional methods heavily reliant on external information, the proposed method employs LLM to discern the shared semantic features of highly impactful papers from a large collection of title-abstract pairs. These semantic features are further utilized to predict the proposed indicator, $\mathrm{{TNCSI}_{SP}}$, which incorporates favorable normalization properties across value, field, and time. To facilitate parameter-efficient fine-tuning of the LLM, we have also meticulously curated a dataset containing over 12,000 entries, each annotated with titles, abstracts, and their corresponding $\mathrm{{TNCSI}_{SP}}$ values. The quantitative results, with an MAE of 0.216 and an NDCG@20 of 0.901, demonstrate that the proposed approach achieves state-of-the-art performance in predicting the impact of newborn articles when compared to several promising methods. Finally, we present a real-world application example for predicting the impact of newborn journal articles to demonstrate its noteworthy practical value. Overall, our findings challenge existing paradigms and propose a shift towards a more content-focused prediction of academic impact, offering new insights for article impact prediction.
\end{quote}
\end{abstract}

\section{Introduction}

% 【what is the field of the work】 
% Recent explorations of LLM-based autonomous agents in automated scientific discovery have highlighted the importance of predicting or evaluating the impact of individual research papers.

The emerging field of article impact prediction is becoming increasingly critical in advancing scientific research. Generally, it focuses on forecasting the potential future citation counts of academic publications by exploiting the external data related to the article~\cite{xia2023review}, such as early citation feature, venue characteristics, and author reputation, \textit{etc}. Unlike traditional bibliometric evaluations that measure established influence, article impact prediction typically encompasses a broader range of applications. Large institutions utilize it for research funding decisions and academic promotions. Individuals may also benefit from impact predictions, which help them efficiently identify cutting-edge articles and remain leading in their fields, especially given the hundreds of daily arXiv submissions across various academic disciplines. 
% 可以额外给出现在投稿量大的背景，比如每天cs中有平均多少稿件发表，数量太大，这个有助于比如effectively/efficiently filter valuable manuscripts given 大量的稿件。

% article impact predictions have been commonly utilized by large institutions for purposes such as research funding or academic promotions. Until recently, article impact prediction has attracted the attention of individual scholars and has been employed in academic activities such as identifying high-quality papers.
% As reported in~\ref{}, 
% In addition to large institutions, individual researchers may also utilize article impact as a signal to identify high-quality papers that are worth reading in their daily academic activities.
% Traditionally, article impact predictions have been commonly utilized by large institutions for purposes such as research funding or academic promotions.
% There are some earlier explorations yielding acceptable results and sufficiently addressing these needs~\cite{}.
% the article's impact reflects whether it is worth being read or not. Thus, 

\begin{figure}[t]
     \centering
     \includegraphics[width=0.99\linewidth]{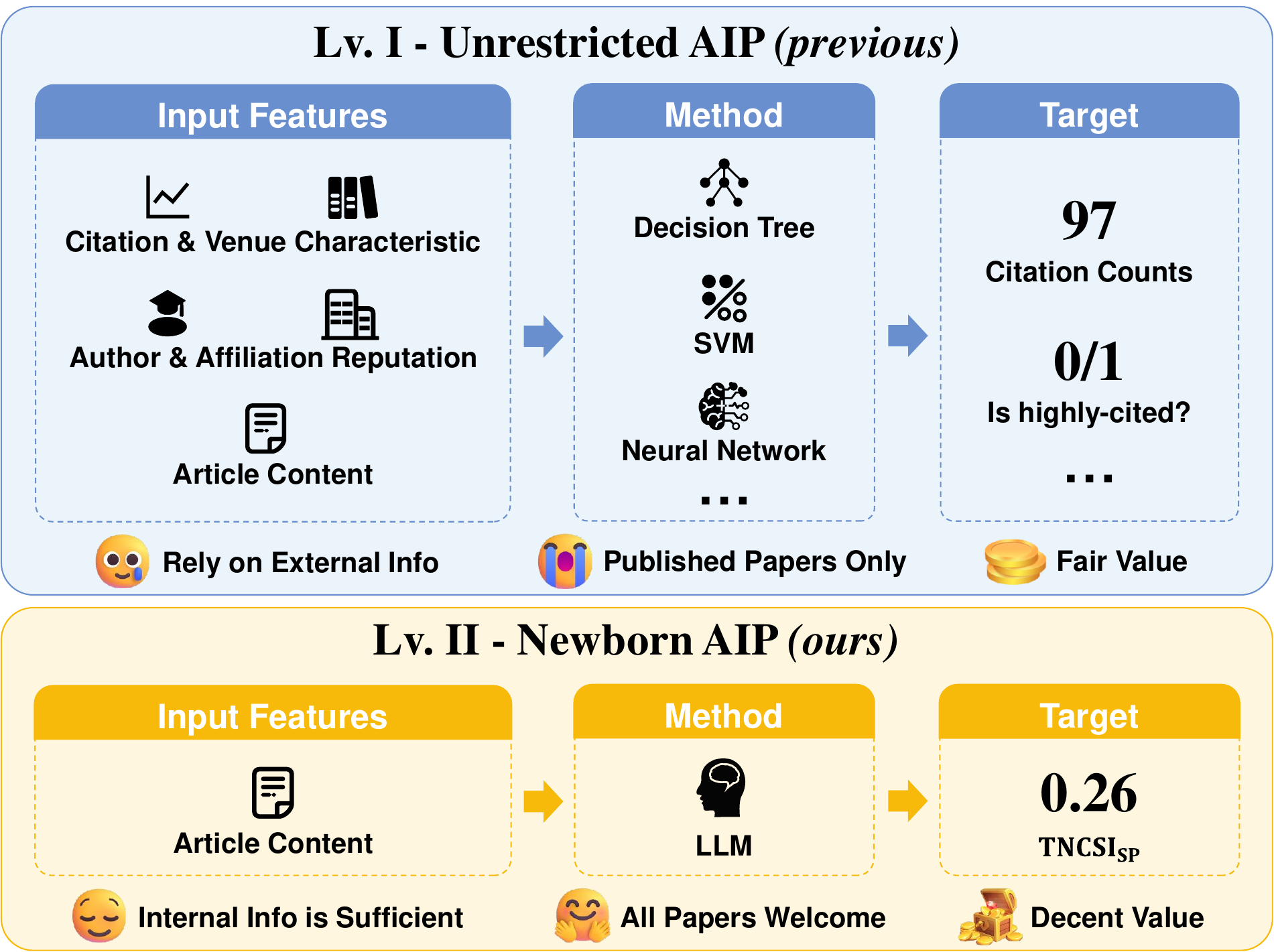}
     \caption{A taxonomy of article impact prediction (AIP): since there are virtually no other Lv. II methods, the ``newborn AIP'' segment represents the proposed approach, which predicts future academic impact in a ``double-blind peer-review'' manner. }
     \label{fig:taxnomy}
 \end{figure}
% 【why is this field important,】
% Unlike impact evaluations that measure established influence, impact predictions usually encompass a broader range of applications. Especially a There is still a growing number of scholars focusing on accurately predicting an article's impact.
Recently, as the field of Large Language Model (LLM) agent-based automated scientific research systems rapidly evolves~\cite{de2023artificial,wang2023scientific,lu2024ai}, article impact prediction has never been more important than it is today. Imitating human experts, these autonomous systems typically start with identifying the most relevant and valuable research literature from extensive academic articles. Only then do these systems extract and synthesize knowledge from the retrieved literature, thereby enabling practical applications such as idea generation~\cite{baek2024researchagent}, and compound discovery~\cite{m2024augmenting}, \textit{etc}. As the saying goes, \textit{You can't make bricks without straw}, article impact prediction has thus become a central component of automated research systems.

% This is because the article impact prediction plays a central role in the paper screening phase of automated scientific research systems. 

However, almost all existing impact prediction approaches rely on external historical data~\cite{Vergoulis2020SimplifyingIP,wang2021prediction,zhao2022utilizing,abbas2023predicting,zhang2024predicting}, which limits the practical value of these methods. Particularly, for those newly uploaded papers on pre-print websites (\textit{e.g.},
arXiv), the absence of historical citation data and publication venue information poses challenges for existing methods to make accurate predictions. In addition, although most academic efforts prefer to predict citation counts, the validity of citation counts themselves remains debatable. As the Leiden Manifesto ~\cite{hicks2015bibliometrics} and the DORA Declaration~\cite{san2018san} indicate, citation count is not well suited for cross-disciplinary comparable evaluations and should not be used as the sole metric for assessing research impact. For example, a paper~\cite{li2023large} with one hundred citations might be unsurprising in the booming field of artificial intelligence, but in a relatively narrow yet equally important field such as paleontology, the paper with one hundred citations~\cite{turner2011paleontology} could be considered a cornerstone. Since automated scientific research systems mostly start from estimating the value of articles, such limitations undoubtedly weaken their ability to gather knowledge from other fields, thereby reducing the efficiency of knowledge synthesis.  % 这段写的很好，两个问题都非常明确的指出来了，1）用历史信息的方法对于新论文不行；2）引用次数不合理。从写作的逻辑上来讲，下面一段就直接给出本文的所提的方式方法来一一解决上面两个问题，针对2）提出新指标；针对1）提出基于新指标对于新论文仅采用内容文本进行LLM微调预测。 然后再下一段再讲我们进一步梳理了3类XXX，然后本文的focus在第三类上。

To address the potential issues of regarding citation counts as the regression target, we first draw inspiration from the design principles of the Topic Normalized Citation Success Index (${\mathrm{TNCSI}}$)~\cite{zhao2024literature} and make tailored improvements to adapt it for predicting the impact of newborn papers across various fields. The improved metric is named $\mathrm{{TNCSI}_{SP}}$, where $SP$ stands for the Same Period, to highlight that the proposed metric is capable of comparing papers across varying time frames. Since key elements such as contribution, novelty, and insights are often reflected in the title and abstract, we contend that the ``\textit{worth}'' of an article can typically be assessed by ``\textit{words}'',  Therefore, we try to regress the $\mathrm{{TNCSI}_{SP}}$ by feeding only the title and abstract to fine-tuning the LLMs for reliable impact predictions.

For better clarity on the current state of impact prediction methods, we summarize and introduce a taxonomy based on the information required for the prediction (See in Fig.~\ref{fig:taxnomy}). The first level is called unrestricted article impact prediction, where predictions are permitted to rely on external historical information, and authors' reputation; this is the level at which most current methods are situated. The second level is named newborn article impact prediction, which particularly emphasizes making predictions about the impact only based on the article itself. This task is similar to a double-blind review process, where the model predicts the future impact without any author and affiliation information, publication details, or early citation data. Such an approach is particularly valuable for screening newly uploaded manuscripts, such as arXiv pre-prints and conference papers, as it may help researchers effectively identify the most promising articles. In this paper, we focus on the most challenging yet most valuable task: newborn article impact prediction.

To summarize, the core contributions of this work are as follows:
\begin{itemize}

\item \textbf{New Task}: We introduce a taxonomy and define a novel task entitled newborn article impact prediction, which aims to accurately predict the scholarly impact of newly published articles without external information.

\item \textbf{New Method}: Tailored improvements have been made to the ${\mathrm{TNCSI}}$, and for the first time, we demonstrate that fine-tuned LLMs are capable of predicting the future impact of newborn articles in a ``double-blind review'' setup.

\item \textbf{New Dataset}: Accordingly, we have constructed and released the TKPD and NAID datasets. They are used to guide ChatGPT in generating topic key phrases and to train state-of-the-art LLMs for accurate article impact predictions, respectively.

\item \textbf{Application}: Finally, we discuss and present an example of the proposed method's application in a real-world scenario, specifically in predicting the impact of journal articles published in 2024, with the hope of inspiring further advancements in the broader research field.
% We highlight the limitations of traditional methods for predicting article impact, particularly their heavy reliance on external information and limited performance on newly published papers. Furthermore,
% Tailored improvements have been made to the TNCSI metric, alongside the exploration of state-of-the-art LLMs to predict the improved metric.  For the first time, we demonstrate that large language models are capable of predicting the future impact of articles with internal information only, thereby providing a new perspective for future research.
% ccordingly, we have constructed and released the TKPD and the NAID dataset. The TKPD dataset utilized for prompt engineering includes over 200 topic key phrase samples annotated by a human expert. NAID dataset comprises titles, abstracts, $\mathrm{{TNCSI}_{SP}}$, and more related information of approximately 12,000 articles published between 2020 and 2022.
 % in the fields of computer vision, natural language processing, and other AI fields
% 指标评价任何一个学术材料的指标也是贡献 which includes  entries of articles across various research fields.

\end{itemize}
Our code framework, datasets, web demo are released at \url{sway.cloud.microsoft/KOH09sPR21Ubojbc}. % 不匿名

% \url{sway.cloud.microsoft/iBwzbJ7vGdhxrNat}. % 匿名
\section{Related Work}

\textbf{Bibliometrics} is a research field that utilizes quantitative analysis and statistical methods to assess the impact of scholarly publications. Typically, bibliometrics can be divided into two major categories: metrics for evaluating journals and metrics for evaluating individual articles. As the Leiden Manifesto~\cite{hicks2015bibliometrics} and the DORA Declaration~\cite{san2018san} recommend, \textit{do not use journal-based metrics to measure the quality of individual research articles}. Therefore, in this paper, we do not intend to use any journal-level bibliometric indicators (such as JIF~\cite{garfield1955citation}) as inputs or prediction targets. Instead, we focus on bibliometric indicators for individual articles. Tab.~\ref{tab:metric_compare} illustrates the differences among them. Although FWCI and RCR are excellent metrics, their non-normalized numerical properties may impair the convergence of neural networks. Proposed by Zhao \textit{et at.}, ${\mathrm{TNCSI}}$~\cite{zhao2024literature} features a clear physical meaning and favorable mathematical properties, representing the probability (ranges between 0 and 1) that an article's impact surpasses that of other articles in the same field. However, ${\mathrm{TNCSI}}$ is initially designed to evaluate review papers across different fields and is therefore not suitable for assessing normal research papers. Furthermore, ${\mathrm{TNCSI}}$ primarily focuses on assessing the existing impact of a review paper and does not normalize the impact of papers published in different years. This may lead to potential unfair comparisons in newborn article impact prediction tasks. Therefore, we propose an improved version to address the limitations of ${\mathrm{TNCSI}}$. More details can be found in the Approach section.

\begin{table}[ht]
    \renewcommand\arraystretch{1.1}
    \begin{center}
        \begin{tabular}{p{0.5\linewidth}p{0.1\linewidth}p{0.1\linewidth}p{0.1\linewidth}}
            \hline
            Bibliometric  &  Value & Field & Time\\ 
            \hline
            \hline
            Cites   & $\times$ & $\times$ & $\times$\\
            % \hline
            FWCI~\cite{colledge2014snowball}   & $\times$ & \checkmark & \checkmark\\
            % \hline
            RCR~\cite{hutchins2016relative}  & $\times$ & \checkmark & \checkmark\\
            % \hline
            TNCSI~\cite{zhao2024literature}  & \checkmark & \checkmark & $\times$\\
            % \hline
            TNCSI$_{\mathrm{SP}}$ (\textit{Ours})   & \checkmark & \checkmark & \checkmark\\
            \hline
        \end{tabular}
    \end{center}
    \caption{Several article-level bibliometrics for evaluating scholar Impact: value, field, and time respectively indicate whether the metric is a value between 0 and 1, whether it allows for cross-field comparisons, and whether it is suitable for the comparison of papers published at different times. These normalizations facilitate the network training.}
    \label{tab:metric_compare}
\end{table}

\noindent \textbf{Article Impact Prediction} approaches typically adopt machine learning methods to forecast the future impact of articles. Most existing methods tend to exploit article statistical features, author characteristics, journal attributes, and historical citation data to aid decision trees, LSTM, MLP, and other machine learning algorithms in making predictions~\cite{fu2008models,wang2011mining,qiu2024early,kousha2024factors}. Ruan \textit{et al}.~\cite{ruan2020predicting} aims to enhance the prediction accuracy of five-year citation counts using a four-layer Back Propagation (BP) neural network by leveraging multiple features related to papers, journals, authors, references, and early citations. Ma \textit{et al.}~\cite{ma2021deep} propose a citation count prediction model that uses early citations and paper semantic features as input and employs Bi-LSTM for final predictions. Another notable citation-based machine learning approach exploits static features and time-dependent citation features to predict potentially excellent papers~\cite{hu2023identifying}. In ABBAS's work~\cite{abbas2023predicting}, an MLP-based method leveraging only external features is proposed to make prediction of future citation counts, achieving a decent performance with an NDCG of 0.95. Zhang et al.~\cite{zhang2024predicting} discover that employing different models for papers in various domains significantly enhances the accuracy of prediction by leveraging early citation data. De~\cite{de2024can} attempts to guide ChatGPT-4 in scoring over 2,000 paper abstracts from multiple perspectives, finding that the scores have Spearman correlation coefficients greater than 0.4 with Mendeley readership, and a correlation of 0.18 with citation counts. To the best of our knowledge, there is currently no method capable of accurately predicting the impact of an article based solely on the internal content.

% Qiu \textit{et al.}~\cite{qiu2024early} used three machine learning algorithms, including multiple linear regression, artificial neural network, and classification\&regression tree to develop a model that predicts the long-term citation counts of academic papers based on author numbers, JIF, first author citations, \textit{etc}. Kousha \textit{et al.}~\cite{qiu2024early,kousha2024factors} used statistical regression to predict citation counts from article metadata such as abstract length and country affiliations. 

% To the best of our knowledge, there are currently no attempts that primarily use textual semantic features to predict the impact of articles.

% citation prediction is one of the most effective and popular way to predict scholar impact. 
% citation count has emerged as one of the most frequently used and direct targets, favored for its conceptual simplicity and ease of access. 

\noindent \textbf{Large Language Models} have demonstrated powerful long-form text modeling capabilities and have been widely applied to various NLP tasks over the past few years, including dialogue systems, machine translation, sentiment analysis, \textit{etc.}~\cite{zhao2023survey, Tu24LLMAPP, jiang2024effectiveness} Many commercial large language models~\cite{openai2022chatgpt,openai2023gpt4,google2024gemini,kimi2024website} are not openly accessible, which prevents us from fine-tuning or instruction-tuning them. Therefore, we turned our attention to several excellent open-access large language models. LLaMA series~\cite{touvron2023llama,meta2024llama3} are advanced language models created by Meta AI, available in multiple versions ranging from 7B to 70B parameters. It demonstrates decent performance on most tasks and has been widely adopted for various applications. Apart from LLaMA, there are several other notable open-source large language models, such as Qwen~\cite{bai2023qwen}, Mistral~\cite{jiang2023mistral}, Falcon~\cite{almazrouei2023falcon}, \textit{etc}. Regardless of the specific large language model, they were originally developed for autoregressive text generation. In this study, we use only the first generated token for numerical regression. Detailed descriptions and comprehensive evaluations of these models will be provided in the Approach and Experiment sections.

% In the past one to two years, we have witnessed the success of large language models, particularly in the domains of text generation, natural language understanding, and conversational systems.

\section{Approach}

\subsection{Tailored Improvement to the TNCSI}
As mentioned in the Related Work section, ${\mathrm{TNCSI}}$ suffers from certain limitations, such as being restricted to evaluating review papers and only taking into account the cumulative impact of articles. We conducted a detailed analysis of its computational process and identified the reasons behind these limitations. 
First, ${\mathrm{TNCSI}}$ requires a predefined prompt template to guide ChatGPT in generating a corresponding review research area from the given title and abstract. The original prompt is specifically designed for review papers rather than normal research papers. Therefore, using their prompt directly on regular papers results in poor performance. Second, ${\mathrm{TNCSI}}$ primarily considers the cumulative impact of an article since its publication. However, for constructing datasets of the article impact prediction task, this approach may lead to potential issues of unfair comparison. Specifically, papers published earlier typically exhibit higher ${\mathrm{TNCSI}}$ values than recently published ones. This could potentially confuse the network's learning process, making LLM difficult to model the relationship between text features and its impact values.

 \begin{figure*}[ht]
     \centering
     \includegraphics[width=1.0\linewidth]{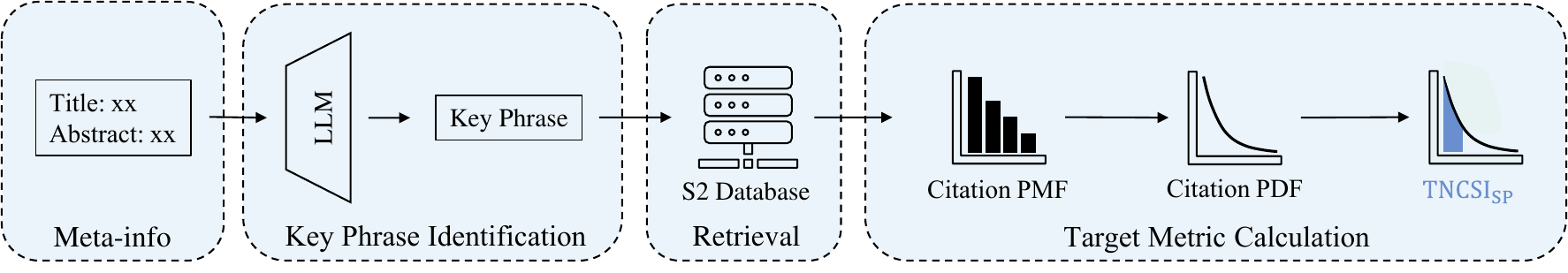}
     \caption{Flowchart for calculating $\mathrm{{TNCSI}_{SP}}$: ${\mathrm{TNCSI}}_{\mathrm{SP}} \in [0, 1]$ represents the probability that a paper's citation count outperforms other papers in the same field and time period. ``S2'' refers to Semantic Scholar.}
     \label{fig:stat_database}
\end{figure*}
Based on the discussion above, we make tailored improvements to the ${\mathrm{TNCSI}}$ and name the improved metric as $\mathrm{{TNCSI}_{SP}}$. Similar to the computational procedure of ${\mathrm{TNCSI}}$, the procedure of the proposed $\mathrm{{TNCSI}_{SP}}$ is divided into three steps. In the first step, a well-designed prompt is utilized to guide ChatGPT (currently refer to gpt-3.5-turbo-0125) to identify the topic key phrase of an article. We have designed and tested a variety of prompt templates for identifying the article key phrase from different perspectives. To further mitigate individual cognitive biases, we enlisted the help of numerous researchers in the prompt creation process. All prompt templates are tested on a human-annotated dataset to evaluate the corresponding performance in the key phrase identification task. The second step involves using ChatGPT-generated key phrases to retrieve 1,000 related papers and their mata-info (\textit{e.g.,} citation counts) from the Semantic Scholar API. Unlike ${\mathrm{TNCSI}}$, which considers citation counts over the entire timeframe, $\mathrm{{TNCSI}_{SP}}$ focuses on the concurrent papers within a 6-month window before and after the publication date.
%solely on the number of citations within a \implus{similar period, e.g., 6-month} window before and after the publication date of the current paper. 
This approach ensures that each paper is compared only to others published within a similar timeframe, thereby minimizing the citation advantage that older papers accumulate due to their extended presence. As a result, this method endows $\mathrm{{TNCSI}_{SP}}$ with the ability to normalize citation impact across different publication times. The final step remains consistent with ${\mathrm{TNCSI}}$. The simplified mathematical expressions are shown as follows:

\begin{equation}
\label{eq_c_k}
    P(X=x) = 
    \begin{array}{ll} 
        \frac{{\text{Count}(p_x)}}{C}.
    \end{array}
\end{equation}

Here, $C = 1000$ refers to the total number of retrieved papers. $\text{Count}(p_x)$ represents the number of paper $p$ with $x$ citations. $P(X=x)$ is a discrete probability distribution that describes the probability of a paper having exactly $x$ citations among the retrieved $C$ papers.

In their work~\cite{zhao2024literature}, $P(X=x)$ has been thoroughly discussed and is shown to follow an exponentially decaying distribution. Therefore, it could be converted into a probability density function using the maximum likelihood estimation. As shown in Eq.~\eqref{eq_TNCSI}, we may derive the final ${\mathrm{TNCSI}}_{\mathrm{SP}} \in [0, 1]$ by calculating the value of the corresponding definite integral:
\begin{equation}
  \label{eq_TNCSI}
    TNCSI_{SP} = \int_0^{cites} \lambda e^{-\lambda x} \,dx, x \geq 0 ,
\end{equation}
\noindent where $cites$ represents the number of citations that the paper being evaluated has received.  % 这里提一下这个指标的数值范围

\subsection{LLM for Newborn Article Impact Prediction}
The autoregressive mechanism of large language models has been well-documented~\cite{zhao2023survey}. Essentially, these decoder-only models generate text in a sequential manner, with each token prediction relying on the context provided by the previous tokens. Such a paradigm allows it to fully leverage unlabeled data for self-supervised learning. 

\begin{figure}[ht]
     \centering
     \includegraphics[width=1.0\linewidth]{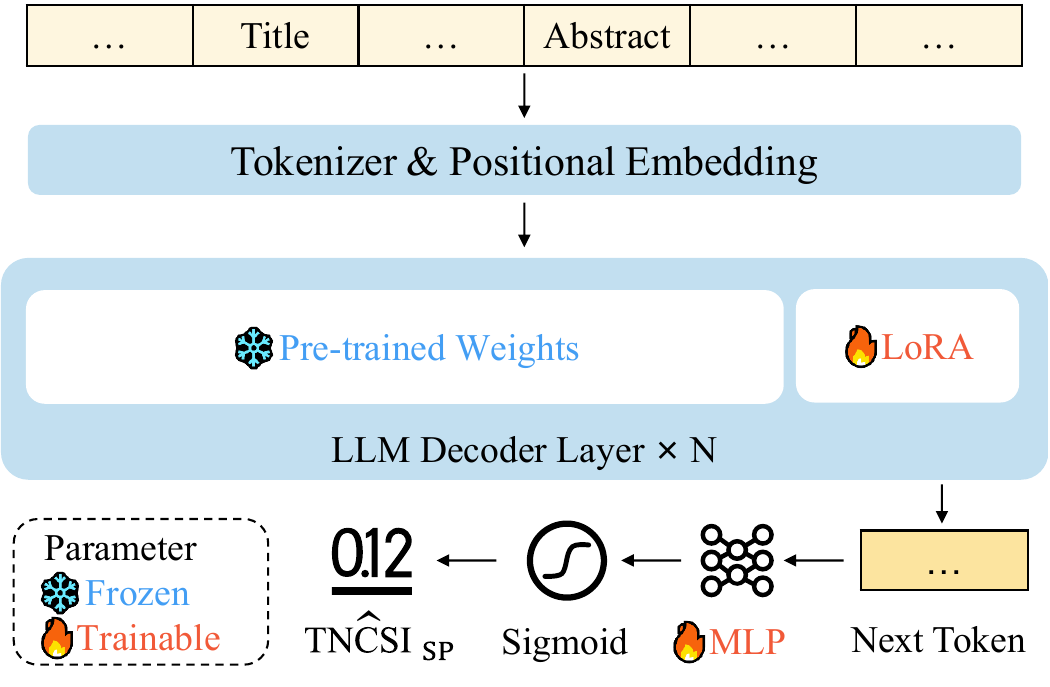}
     \caption{LLM as scholar impact predictor: overall framework of the proposed approach. Only the Next Token (first generated token) is used to regress the ${\mathrm{TNCSI}}_{\mathrm{SP}}$. 
     % 这里图里面不用这么多decoder layer，浪费空间，2个就行，中间点点点
     }
     \label{fig:database_construction}
\end{figure}
In this paper, we maintain the autoregressive generation scheme of the large language model unchanged. However, unlike conventional text generation, we focus solely on the first token that the model generates autoregressively in response to user input. Specifically, assume the current input sequence is $\{w_1, w_2, \ldots, w_t\}$. The relationship between the LLM and the generation of the next token $w_{t+1}$ can be expressed as:

% \begin{equation}
% \label{eq_llm_complicate}
% w_{t+1} = \arg\max_{w} \, P(w \mid w_1, w_2, \ldots, w_t)
%  \end{equation}
% Let the large language model be denoted as $LLM(\cdot)$, from which we can derive an alternative, concise form of Eq.~\ref{eq_llm_complicate}.
\begin{equation}
w_{t+1} = \text{LLM}(w_1, w_2, \ldots, w_t),
\end{equation}

\noindent where \( \text{LLM}(\cdot) \) represents a large language model that predicts the next token in the sequence based on the input token sequence $\{w_1, w_2, \ldots, w_t\}$.

 To facilitate the LLM's prediction of a single numerical value, we employ a simple multi-layer perceptron (MLP) to transform $w_{t+1} \in \mathbb{R}^{B \times 1 \times D}$ into a real number $v \in \mathbb{R}$. Then, the value $v$ is fed to a Sigmoid function, resulting in the predicted $\mathrm{\hat{TNCSI}}_{\mathrm{SP}} \in [0, 1]$. Here, $B$ represents the batch size, $D$ is the dimension, and $\mathbb{R}$ denotes the set of real numbers. The process can be represented by the following equations:
 
% \begin{equation}
% v = 
% \end{equation}

\begin{equation}
\hat{TNCSI}_{SP} = \sigma(\text{MLP}(w_{t+1})),
\end{equation} 

% \noindent where  \( v \) is computed using a multi-layer perceptron applied to \( w_{t+1} \), and   \( \hat{TNCSI}_{SP} \) is the result of applying the Sigmoid function \( \sigma \) to \( v \).

\noindent where $\mathrm{\hat{TNCSI}}_{\mathrm{SP}}$ is computed by passing $w_{t+1}$ through an MLP followed by a Sigmoid function $\sigma$.

Finally, we aim to minimize the mean square error (MSE) loss to align the predicted output $\mathrm{\hat{TNCSI}}_{\mathrm{SP}}$ to the $\mathrm{{TNCSI}_{SP}}$ obtained from previous statistical calculations.

% In practice, due to the vast number of parameters in large language models, training each parameter in the model requires significant computational resources that are difficult to imagine. 
In practice, the immense number of parameters in large language models requires substantial computational resources for training, which exceeds our practical capacity. Therefore, we adopted low-rank matrix decomposition (LoRA)~\cite{hu2021lora} and model quantization techniques~\cite{dettmers2022gpt3} to reduce computational resource consumption and accelerate network training and inference processes. We recommend readers refer to the original papers for further details. 

\subsection{Datasets Construction}
\label{sec:dataset}
We have constructed a total of two datasets, the Topic Key Phrase Dataset (TKPD) and the Normalized Article Influence Dataset (NAID). Each of these datasets serves different purposes, which will be described in more detail below.

\noindent \textbf{Topic Key Phrase Dataset}: 
 TKPD includes 251 entries encompassing titles, abstracts, and core task or field names of random articles across various fields in artificial intelligence. To mitigate subjectivity in our study and ensure consistent annotations, we employ manual labeling of key phrases by a seasoned AI researcher and enlist three additional researchers to double-check the annotations. Due to the specialized knowledge required for data annotation, this paper is precluded from annotating articles from non-AI fields. Nevertheless, we believe that the inconsistencies between the different subfields within the AI field are sufficient to simulate the differences between the distinct disciplines.

% To avoid the shadow effect, we adhere to manual annotation to ensure the validity of prompt engineering. It is also important to note that labeling key phrases based on titles and abstracts involves a certain degree of subjectivity. There, the labeling procedure is performed by a single senior researcher specializing in the AI field to ensure consistency in the annotations. As mentioned in the Approach section, we have tested the effectiveness of various prompts on the TKPD.

% Consequently, this paper is precluded from annotating articles in non-AI fields, thereby hindering analysis of articles in other domains.

\noindent \textbf{Normalized Article Impact Dataset}: 
NAID is used to train LLMs to predict the impact of articles. It comprises the title, abstract, and the corresponding $\mathrm{{TNCSI}_{SP}}$, \textit{etc}. The NAID consists of over 12,000 data entries from various AI fields, excluding survey papers, and includes papers with category IDs "cs.CV", "cs.CL", and "cs.AI" uploaded to arXiv between 2020 and 2022. In particular, the "cs.AI" category spans a broad range of disciplines such as mathematics, physics, and cognitive science, thereby extending the training data beyond the AI domain. NAID is a uniformly distributed dataset, meaning that the sources of the papers, the original publication year of the papers, and the corresponding $\mathrm{{TNCSI}_{SP}}$ values are evenly distributed.

% This uniformity aids in accelerating network convergence. Specifically, we denote the NAID subset composed of papers published in a particular year as NAID-year, e.g., NAID-2021, indicating that all data in this subset correspond to papers published in 2021.where pre-annotating with Prompt A might lead to artificially inflated performance in testing

% Given the significantly larger scale of our database compared to that constructed by Zhao et al., using a single-threaded approach to build the dataset proves exceedingly slow. Consequently, we have implemented substantial engineering optimizations, enabling the current codebase to construct the dataset asynchronously. This approach maximizes dataset construction speed without violating the Semantic Scholar protocol.  Subsequently, we randomly selected about 3,000 articles from each year as the sample data constituting the NAID.
\section{Experiments}

% 

% 一定要说明由于标注数据需要专业领域知识，我们无法标注非AI领域的数据集。因此，本文局限于AI领域。但需要指出AI领域内，不同子领域差异也很大，足够说明问题
\subsection{Metrics}

\noindent \textbf{Mean Absolute Error}: MAE is employed to assess the prediction accuracy. It is a measure used to evaluate the discrepancy between the predicted value and the ground truth $y_i$, which is defined as follows:
\begin{equation}
    \text{MAE} = \frac{1}{n} \sum_{i=1}^{n} |y_i - \hat{y}_i|,
\end{equation}
where $n$ represents the number of samples in the test set,$y_i$ denotes the actual output (\textit{e.g.}, $\mathrm{{TNCSI}_{SP}}$), $\hat{y}_i$ stands for the predicted value (\textit{e.g.}, $\hat TNCSI_{SP}$), $|y_i - \hat{y}_i|$ is the absolute difference between the actual and predicted values. Generally, a lower MAE indicates a higher accuracy of the model's predictions.

\noindent \textbf{Normalized Discounted Cumulative Gain}~\cite{jarvelin2000ir}:
% To better illustrate the performance of various approaches, a new impact rank task has been devised. The process begins by ranking the \( l \) papers within the test set according to their real \( TNCSI_{SP} \) values, forming an ordered sequence \( Seq_{GT} = (p_1, p_2, \ldots, p_l) \). Subsequently, these papers are ranked based on their predicted \( TNCSI_{SP} \) values to produce another sequence, \( Seq_{Pred} \). Each \( p \) in these sequences represents a specific article entry.
 NDCG is another metric to evaluate the effectiveness of the prediction. NDCG, originally developed for recommendation systems to measure the gain of a document based on its position in the recommended list, is calculated as follows:
\begin{equation}
    \label{eq:NDCG}
    \text{NDCG@K} = \frac{\text{DCG@K}}{\text{IDCG@K}},
\end{equation}
where DCG@K = $\sum_{i=1}^{K} (2^{\hat{y}_i}-1)/\log_2(i+1)$, and IDCG@K = $\sum_{i=1}^{K} (2^{y_i}-1)/\log_2(i+1)$. $K=20$ represents the position cutoff for the recommended list. NDCG is a metric that ranges from 0 to 1, with scores closer to 1 reflecting that more influential documents are ranked higher, signifying better performance.

% where \text{DCG@K} and \text{IDCG@K} are computed as:
% \begin{equation}
% \begin{aligned}
%     \text{DCG@K} &= \sum_{i=1}^{K} \frac{2^{\hat{y}_i} - 1}{\log_2(i+1)}, \\
%     \text{IDCG@K} &= \max \left( \sum_{i=1}^{K} \frac{2^{{y}_i} - 1}{\log_2(i+1)} \right).
% \end{aligned}
% \end{equation}

% \noindent \textbf{Kendall's Tau coefficient} is a non-parametric statistic used to measure the ordinal association between two sequences. It evaluates the strength of the relationship by considering the concordance and discordance of all possible pairs of elements $(i, j)$ in the sequences. The formula for Kendall's Tau is defined as:

% \begin{equation}
% \tau = \frac{2(N_c - N_d)}{n(n-1)}
% \end{equation}

% \noindent where $N_c$ is the number of concordant pairs, $N_d$ is the number of discordant pairs, and $n$ is the number of elements in $Seq_{GT}$.

% \noindent  \textbf{Spearman's Rank Correlation Coefficient} is another non-parametric measure that quantifies the strength and direction of the monotonic relationship between two ranked variables. It evaluates how well the relationship between two variables can be described using a monotonic function. The formula for Spearman's Rank Correlation is given by:

% \begin{equation}
% \rho = 1 - \frac{6 \sum d_i^2}{n(n^2 - 1)}
% \end{equation}

% \noindent where $d_i$ represents the difference between the ranks of each pair of elements in the $Seq_{GT}$ and $Seq_{Pred}$.
 
% The values of Kendall's Tau and Spearman's Rank Correlation range between [-1, 1]. A value approaching 1 indicates that the predicted order is more similar to the true order.

\noindent \textbf{Normalized Edit Distance}~\cite{yujian2007normalized}: NED is a metric to measure the similarity between two strings by normalizing the edit distance by the length of the longer string. It is defined as: 
\begin{equation}
\text{NED}(A, B) = \frac{\text{ED}(A, B)}{\max(|A|, |B|)},
\end{equation}

\noindent where \(\text{ED}(A, B)\) is the edit distance between strings \(A\) and \(B\), and \(\max(|A|, |B|)\) is the length of the longer string. The lower the NED value, the more similar the two strings are.

\subsection{Comparison with Previous Methods}
We employ NDCG to assess the effectiveness of different methods in identifying high-impact papers, given their varying prediction targets. To ensure fairness, we exclude the external data relied upon by previous approaches. Reproduction details are provided in the Appendix.

% Due to the different prediction targets of various methods, we employ NDCG to evaluate the effectiveness of these methods in identifying potentially high-impact papers. As previously mentioned, nearly all methods heavily rely on external data from the papers. Therefore, to ensure a fair comparison, we removed the external information relied upon by previous methods. Details of the reproduction process are provided in the Appendix.

% As previously mentioned, the article impact prediction methods at different levels rely on varying data and exhibit different task complexities. Directly comparing all methods may lead to potential unfairness. Therefore, the last column of the table presents the results reproduced using the criteria of the third-level newborn article prediction, \textit{i. e.}, removing external journals, citations, authors, and affiliations features while making predictions.It is noted that some articles fail to report the NDCG values for their methods. This subsection compares the proposed method with previous state-of-the-art methods. 

% As previously mentioned, nearly all methods heavily rely on external data from the papers. To adapt to the tasks targeting newly published papers, we have omitted some additional information for certain methods, such as the publishing journal and early citation details, \textit{etc}.

\begin{table*}[ht]
    
    \begin{center}
        \begin{tabular}{p{0.32\linewidth}p{0.07\linewidth}p{0.3\linewidth}p{0.1\linewidth}p{0.08\linewidth}}
            \hline
             Methods & Ori. Lv. & Input Feature for Fair Comparison& Target  & NDCG ↑  \\
            \hline
            \hline
            MLP-based~\cite{ruan2020predicting}& I& paper length, reference numbers, \textit{etc.}&Cites  & 0.147  \\
            % (论文长度): (标题长度):(标题中的标点):(参考文献数量)(参考文献年龄) MLP
            %\hline
            LSTM-based~\cite{ma2021deep} & I& title, and abstract & Cites  & 0.196 \\
            % 只给摘要LSTM
            %\hline
   
            Model Ensemble~\cite{zhang2024predicting}& I & ~\cite{ruan2020predicting} + research filed & Cites  & 0.201  \\ 
            % 根据cs.XX用多个MLP分类。
             MLP-based~\cite{hu2023identifying} & I & the same to~\cite{ruan2020predicting}  &  Is Top 5\%  & 0.464  \\ 
            % 关键词数量 (Number of keywords)摘要长度 (Abstract length)论文长度 (Paper length)参考文献数量 (Number of references)
            % MLP-based & I & the same to~\cite{ruan2020predicting}  & TNCSI$_\mathrm{\text{SP}}$  & x  \\ 
            % LSTM-based~\cite{ma2021deep} & I& the same to~\cite{ruan2020predicting} & TNCSI$_\mathrm{\text{SP}}$  & x \\
            
            \hline
            ChatGPT-generated~\cite{de2024can} & II & title, and abstract & Score  & 0.597   \\
            %\hline
            % LLaMA-3-Generated (\textit{ours})& II & title, and abstract & TNCSI  & 0.649 \\
            LLaMA-3-generated & II & title, and abstract & TNCSI$_\mathrm{\text{SP}}$  & 0.674 \\
            Fine-tuned LLaMA-3-based & II& title, and abstract & Cites  & 0.403 \\
            Fine-tuned LLaMA-3-based & II& title, and abstract & FWCI  & 0.594 \\
            Fine-tuned LLaMA-3-based (\textit{ours}) & II& title, and abstract & TNCSI$_\mathrm{\text{SP}}$  & \textbf{0.901} \\
            \hline

        \end{tabular}
    \end{center}
\caption{Comparison with previous approaches: the proposed method achieves notable advantages among others. An upward arrow indicates that a higher value is better, and vice versa. Bold font denotes the best performance among all methods. `Ori. Lv.' refers to the taxonomy level of the original study, while `Target' denotes the predicted target type in the corresponding research.  See the Appendix for more reproduction details.}

    \label{tab:comp_various_methods}
\end{table*}
% 把Reported去掉，增加每个工作用了哪些信息的说明 \rewrite{Check:}
Tab.~\ref{tab:comp_various_methods} clearly illustrates the performance differences between our proposed method and previous methods in identifying potentially high-impact papers. The proposed method demonstrates a notable superiority in the newborn article impact prediction setting compared to earlier representative works. Most level I methods underperform without external information. For example, the LSTM-based method~\cite{ma2021deep} reports an NDCG of 0.84 when leveraging external information, but its performance drops significantly to 0.196 when relying solely on the title and abstract, suggesting a limited capacity to effectively map semantic features to the target $\mathrm{{TNCSI}_{SP}}$. The less favorable performance of ChatGPT-generated and LLaMA-3-generated approaches in identifying high-impact papers suggests that zero-shot LLM-generated approaches still require further exploration. In summary, we believe the significant performance improvements of the proposed approach could be credited to LLaMA-3's extensive foundational knowledge and the incorporation of the $\mathrm{{TNCSI}_{SP}}$ metric during fine-tuning, which enhances its ability to identify impactful semantic features across various domains and time periods.

% Even when compared to the less challenging Level I tasks~\cite{ma2021deep}, our method still exhibits competitive advantages. Although our method falls short of surpassing the work of Abbas ~\textit{et al.}~\cite{abbas2023predicting}, their approach relies entirely on external data, which makes it unsuitable for predicting newborn articles. Surprisingly, ChatGPT performs poorly in identifying high-impact papers, suggesting that zero-shot-based scoring approaches still require further exploration.

\subsection{Performance of Various LLMs}
% Before reporting the performance, we'd like to make a statement. Though the main purpose of the paper is to explore the effectiveness of LLM on the task of article impact prediction, we are also interested in comparing the proposed method with previous approaches. Unfortunately, none of these papers (including papers not being referenced) have open-source code. And due to manpower and time constraints, we are unable to reproduce their experiments. Therefore, this paper is unable to report the performance difference between the proposed method and the previous methods.
As the central task of this paper, we comprehensively evaluate the performance of various LLMs on the NAID test set. As shown in Tab.~\ref{tab:various_LLM_compare}, LLaMA-3-8B achieves the best overall performance. Interestingly,  we observe that MAE and NDCG are not always inversely correlated; for example, while Falcon achieves a lower MAE than Phi-3, its NDCG is slightly lower. This suggests that Falcon is more accurate in predicting lower-impact papers but less effective for high-impact ones. Since our primary focus is on identifying high-impact papers, a higher NDCG is generally more advantageous than a lower MAE in this scenario.
% As the core task of this paper, we have extensively measured the performance of various LLMs on the NAID test set. The data in Tab.~\ref{tab:various_LLM_compare} shows that LLaMA-3-8B achieved the best performance. Additionally, we observe that MAE values are not always negatively correlated with NDCG values. A lower MAE might not necessarily lead to a higher NDCG. Taking Falcon and Phi-3 as an example, we find that although Falcon achieves a better MAE, its NDCG is slightly reduced when compared to the Phi-3. This may indicate that the model is more accurate in predicting papers with lower influence but less effective in predicting those with higher influence. Considering that our primary focus is to identify high-impact papers, a higher NDCG value is generally more advantageous than a lower MAE in such cases.

%most scenarios.
% It is also worth noting that despite the only 3.8 billion parameters, Phi3 still achieves acceptable performance, making it more suitable for deployment on devices with limited computing power.

\begin{table}[ht]

    \renewcommand\arraystretch{1.1}

    \begin{center}
    
        \begin{tabular}{cccccc}
            \hline
            LLMs & Size ↓ & MAE ↓ & NDCG ↑ & Memory ↓ \\
            \hline
            \hline
            Phi-3& \textbf{3.8B}& 0.226 & 0.742 & \textbf{6.2GB}  \\
            %\hline
             Falcon& 7B& 0.231 & 0.740 & 8.9GB\\
             %\hline
             Qwen-2 & 7B  & 0.223 & 0.774 & 12.6GB  \\
             %\hline
            Mistral& 7B & 0.220 & 0.850 & 15.4GB  \\
             %\hline

             LLaMA-3 & 8B & \textbf{0.216}  &  \textbf{0.901} & 9.4GB  \\

            \hline
        \end{tabular}
    
    \end{center}
    \caption{Performance comparison of different LLMs on the NAID test set: ``Memory'' stands for the minimum memory usage during inference.}
    \label{tab:various_LLM_compare}
\end{table}

\begin{figure}[ht]
     \centering
     \includegraphics[width=1.0\linewidth]{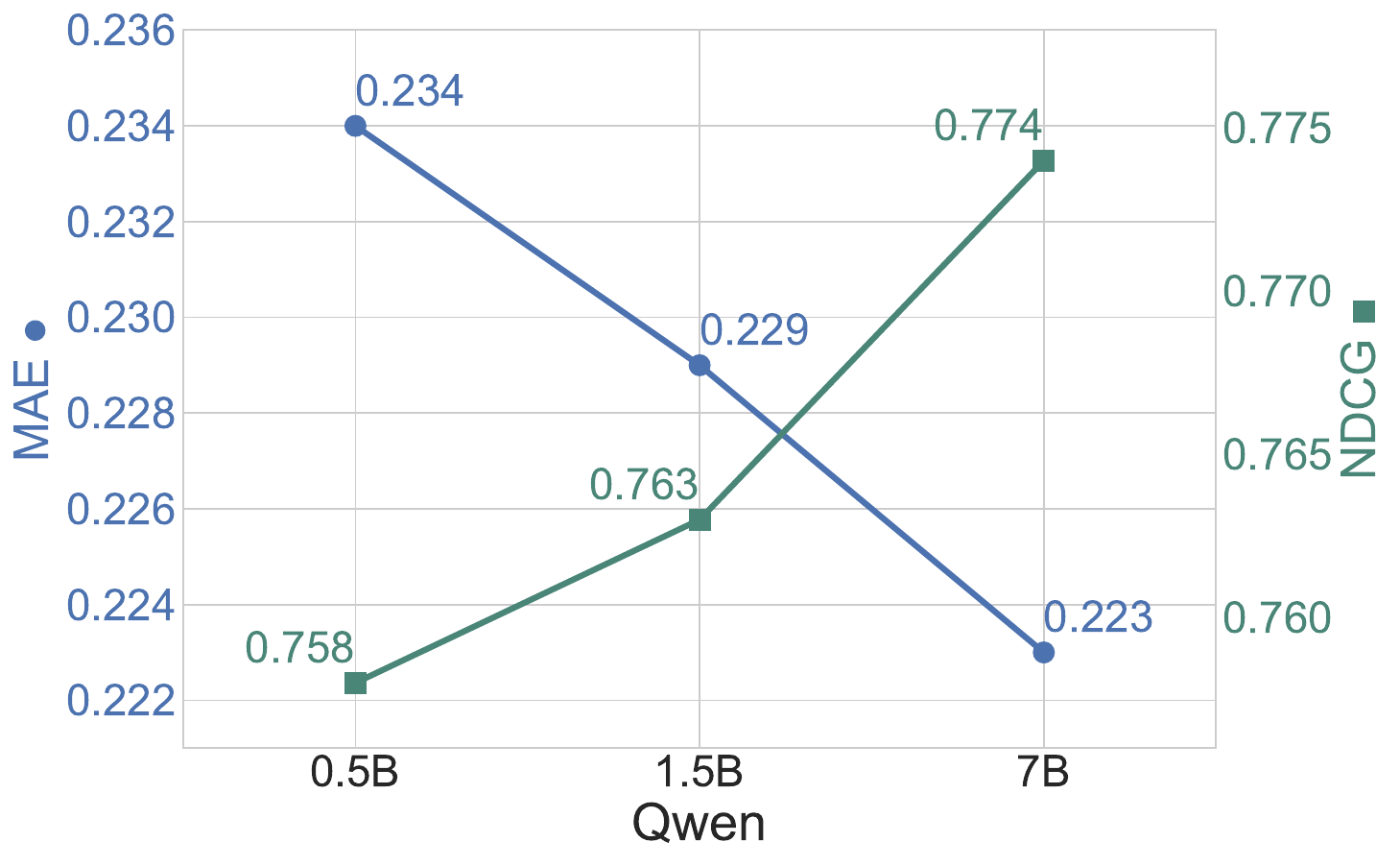}
     \caption{The impact of various model parameters on performance: the larger the number of model parameters, the better the performance.}
     \label{fig:size_vs_performance}
\end{figure}

The Qwen family is selected to further explore the effects of model size on the performance of article impact prediction tasks. Compared to the LLaMA series, the Qwen series features more official models with smaller parameter sizes, specifically 0.5B, 1.5B, and 7B. We train each of these models on the NAID train set, and the test results are illustrated in Fig.~\ref{fig:size_vs_performance}. It can be observed that as the model parameter size increases, the performance correspondingly improves.

\subsection{Effectiveness of Prompt Engineering}

This paper conducts prompt engineering on two tasks: identifying the topic key phrase for calculating $\mathrm{{TNCSI}_{SP}}$, and guiding the LLM to make predictions.

\noindent \textbf{For Identifying Key Phrase}:
We conduct numerous experiments to test the performance of different models and prompts on the TKPD. Tab.~\ref{tab:tkpd_PE} reports the NED for 3 representative prompts on the TKPD. Ultimately, we employ the prompt template from the last row, along with gpt-3.5-turbo-0125, to generate topic key phrases. Extended experimental records can be found in the Supplementary Material. % 别 project page, 别人没有义务看你的project page，直接写 at the Supplementary Materials and our project page. 补充材料里面都写上。

\begin{table}[h]
    \begin{center}
        \begin{tabular}{p{0.75\linewidth}p{0.1\linewidth}}
            \hline
             User Prompt Template & NED↓ \\
            \hline
            \hline
             Identify the research field from the given title and abstract. You MUST respond with the keyword ONLY in this format: xxx  & 0.30 \\
            \hline
            Based on the title and abstract, determine the main area of study for the paper, focusing on a keyword that accurately represents the field. You MUST respond with the keyword ONLY in this format: xxx.  & 0.29 \\
            \hline
            Given the title and abstract below, determine the specific research field by focusing on the main application area and the key technology. You MUST respond with the keyword ONLY in this format: xxx.  & \textbf{0.26} \\
            \hline
        \end{tabular}
    \end{center}
    \caption{Comparison of various user prompts for identifying topic key phrase.}
    \label{tab:tkpd_PE}
\end{table}

\noindent \textbf{For Guiding LLM}: 
As shown in Tab.~\ref{tab:llm_PE}, We test several prompt templates to wrap the title and abstract before inputting them into the fine-tuned LLM. Despite PEFT, variations in prompt templates affect performance; more detailed descriptions often lead to better results. However, overly detailed prompts may also cause a slight NDCG decrease.

\begin{table}[h]

    \begin{center}
        \begin{tabular}{p{0.75\linewidth}p{0.12\linewidth}}
            \hline
             Prompt Template & NDCG↑\\
            \hline
            \hline
              Title: \{title\} \textbackslash n Abstract: \{abstract\}. & 0.849  \\
            \hline

            Given the provided title and abstract, predict the future normalized academic impact on a scale from 0 (lowest impact) to 1 (highest impact). You may consider factors such as the language clarity, novelty of the research, or the claim of state-of-the-art, etc. Title: \{title\}\textbackslash nAbstract: \{abstract\}  & 0.869  \\
            \hline
             Given a certain paper entitled \{title\}, and its abstract: \{abstract\}. Predict its normalized scholar impact:    & 0.889  \\
            \hline
            Given a certain paper entitled \{title\}, and its abstract: \{abstract\}. Predict its normalized scholar impact (between 0 and 1):    & \textbf{0.901} \\

            \hline

        \end{tabular}
    \end{center}
    \caption{Comparison of various prompts for guiding LLMs to predict the future impact. }
    \label{tab:llm_PE}
\end{table}

\subsection{Comparative Analysis of the TNCSI$_{\mathrm{SP}}$}

\begin{figure}[t]
     \centering
     \includegraphics[width=1.0\linewidth]{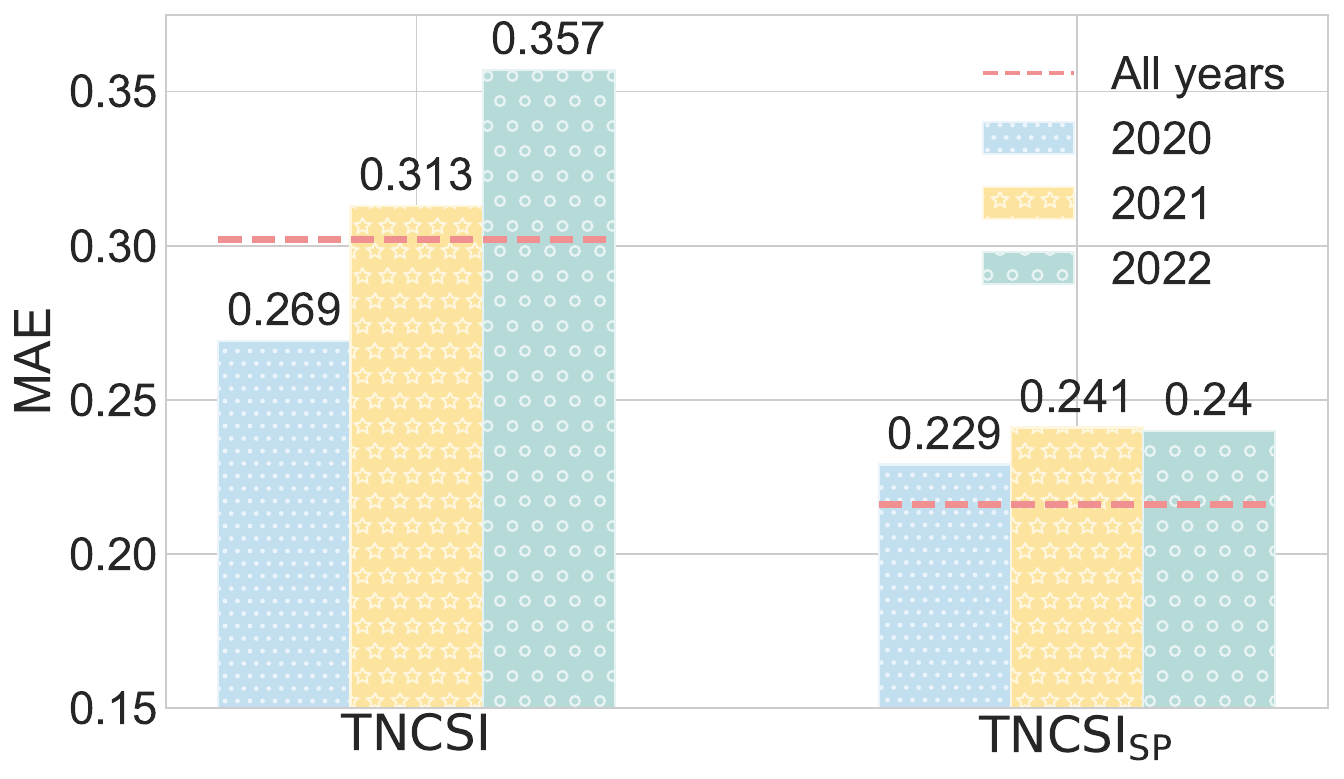}
     \caption{Impact of different prediction targets on performance: $\mathrm{{TNCSI}_{SP}}$ demonstrates superior performance over ${\mathrm{TNCSI}}$ with training data from different years.}
     \label{fig:vary_year}
 \end{figure}

% We PEFT the LLaMA2 (which has only been trained on the data before 2022.9) on the XXX-2023 dataset to show the ... resist temporal cumulative bias and
We have trained LLaMA-3 on articles from different years and with various regression targets to demonstrate the superiority of the proposed $\mathrm{{TNCSI}_{SP}}$. As shown in Fig.~\ref{fig:vary_year}, when targeting the improved $\mathrm{{TNCSI}_{SP}}$, the model provides more stable predictions regarding articles from different years. Table 6 further demonstrates the generalizability of TNCSI$_\mathrm{\text{SP}}$. It enables various types of models to better resist the bias accumulated over time. This suggests that $\mathrm{{TNCSI}_{SP}}$ empowers the model to identify semantic features shared by high-impact articles across different years, thereby achieving significant improvements in overall task performance.

% Tab.~\ref{tab:target_compare} further

% As previously mentioned, since the original TNCSI metric focuses on the cumulative impact, it proves inadequate for predicting the impact of newborn articles (see in Tab.~\ref{fig:vary_year}).

% We utilize LLaMA3 to predict the TNCSI and $\mathrm{{TNCSI}_{SP}}$ values for different years, aiming to demonstrate the superior characteristics of $\mathrm{{TNCSI}_{SP}}$.

\begin{table}[ht]
    \renewcommand\arraystretch{1.1}
    \begin{center}
        \begin{tabular}{lll}
            \hline
            Methods & Target & NDCG ↑ \\
            \hline
            \hline
            % MLP-based & Cites & 0.742   \\
            MLP-based & $\mathrm{\text{TNCSI}}$ & 0.464   \\
            MLP-based & TNCSI$_\mathrm{\text{SP}}$ & 0.634   \\
            \hline
            % LSTM-based & Cites & 0.742   \\
            LSTM-based & $\mathrm{\text{TNCSI}}$ & 0.373   \\
            LSTM-based & TNCSI$_\mathrm{\text{SP}}$ & 0.646   \\
            \hline
            % Fine-tuned LLama-3-based  & Cites  &  \textbf{0.901}   \\
            Fine-tuned LLaMA-3-based   & $\mathrm{\text{TNCSI}}$ & 0.865   \\
            %\hline
            Fine-tuned LLaMA-3-based & TNCSI$_\mathrm{\text{SP}}$ & \textbf{0.901}   \\
            %\hline

            \hline
        \end{tabular}
    
    \end{center}
    \caption{Performance comparison using the TNCSI$_\mathrm{\text{SP}}$ metric: all methods show improvements when targeting TNCSI$_\mathrm{\text{SP}}$. The input is consistent with Tab.~\ref{tab:comp_various_methods}.}
    % \caption{Performance comparison showing the improvement across all methods using the TNCSI$_\mathrm{\text{SP}}$ metric: input data of each method is consistent with Tab.~\ref{tab:comp_various_methods}.}

    \label{tab:target_compare}
\end{table}

\section{Applications}
% We illustrate its utility through three examples: polishing the title and abstract for individual researchers, indicating acceptance in the peer-review procedure, and identifying promising articles for the automated scientific research system. Each example highlights the potential of our approach to enhance various aspects of the research and publication process.This section demonstrates the practical applications of our method for predicting the impact of newborn articles.  JCR quartiles categorize academic journals into four tiers, each representing 25\% intervals based on their impact factors, to reflect their relative influence within a specific field.some scholars recommend evaluating journals based solely on the top 25\% most-highly cited papers~\cite{leydesdorff2012alternatives}. Taking these considerations into account,
In this section, we present an intriguing example, journal average impact prediction, to further demonstrate the effectiveness of our method in real-world applications. 

\begin{figure}[ht]
     \centering
     \includegraphics[width=1.0\linewidth]{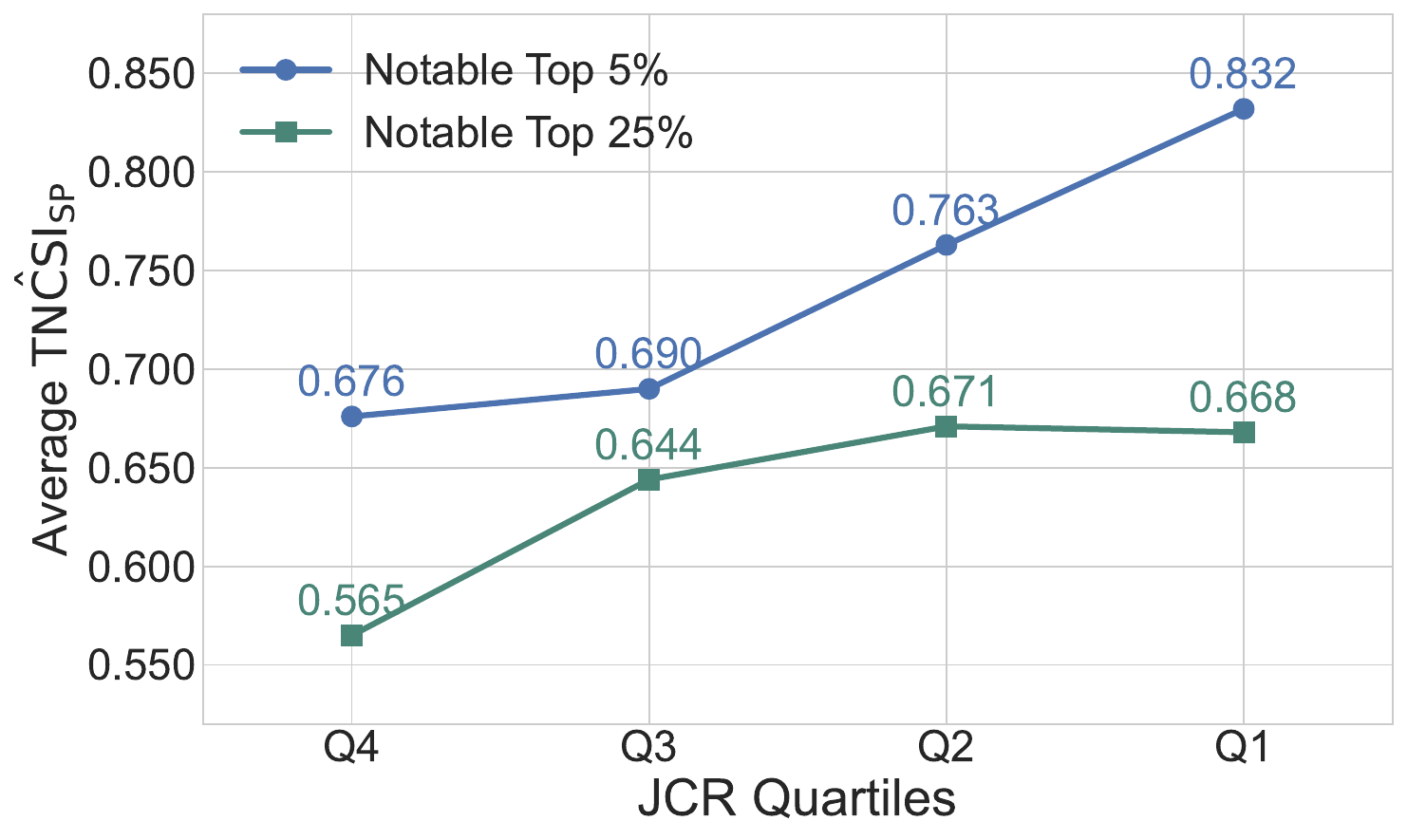}
     \caption{Predicted $\mathrm{{TNCSI}_{SP}}$ values for journals in different JCR quartiles: higher quartiles show higher predicted values. To avoid potential conflicts of interest, we denote Q1, Q2, Q3, and Q4 to represent articles from journals in JCR quartiles 1, 2, 3, and 4, respectively.}
     \label{fig:journal_sets}
 \end{figure}

Theoretically, journals in different quartiles are expected to exhibit varying average impacts. Therefore, we guide the LLaMA-3 to predict the average $\mathrm{{TNCSI}_{SP}}$ of articles published in 2024 across several journals from different JCR quartiles within the field of computer science. Since \mbox{LLaMA-3's} training data only extends up to early 2023, it is highly unlikely that the model has encountered these articles, significantly reducing the risk of data leakage. It is also worth noting that a journal's impact factor would be significantly influenced by a small number of highly cited papers~\cite{lei2020should,leydesdorff2012alternatives}. To this end, we analyzed over 500 randomly selected articles from various journals across different quartiles for impact prediction, focusing on the average predicted $\mathrm{{TNCSI}_{SP}}$ of the top 5\% and 25\% of notable papers within each quartile. In Fig.~\ref{fig:journal_sets}, we observe a clear positive correlation between the predicted impact of the notable top 5\% of articles and their respective quartiles. Although the predicted impact of the top 25\% of articles in the Q2 quartile is slightly higher than that of Q1, it is still considered a reasonable phenomenon. 

Beyond journal impact prediction, our system holds promise for a variety of other real-world applications. For instance, given the vast number of daily pre-print submissions, the proposed approach may also help efficiently identify high-quality research worth closer examination. It may significantly reduce the time researchers spend reviewing large volumes of papers on arXiv, thereby enhancing overall research efficiency.
% Additionally, all selected articles were published between January and August 2024, which makes it highly unlikely for them to appear in the training data of LLaMA-3, thereby minimizing the risk of data leakage.

% just As stated in the Leiden Manifesto~\cite{hicks2015bibliometrics}: \textit{``Do not use journal-based metrics, such as Journal Impact Factors, as a surrogate measure of the quality of individual research articles.''}

% To avoid potential conflicts of interest, we denote the sets A, B, C, and D to represent articles from journals in JCR quartiles 1, 2, 3, and 4, respectively. Semantic Scholar API is utilized to retrieve articles from these journals published between January and August 2024. Since the pre-training data for LLama3 predates March 2023, it is highly unlikely that articles from this period are included in the LLama-3 training data, thus minimizing the risk of data leakage.

% Due to copyright restrictions, we are unable to obtain the PDFs of these journal articles. Therefore, in this application, we provided the LLama-3 with only the title and abstract.
% Predicted $\mathrm{{TNCSI}_{SP}}$ values for journals in different JCR quartiles: Higher quartiles show higher predicted values.

\section{Conclusion}

In this paper, we demonstrate the potential of LLMs for predicting the tailored $\mathrm{{TNCSI}_{SP}}$ of newborn papers with titles and abstracts only. The NAID dataset, comprising over 12,000 entries, is constructed and utilized to fine-tune various advanced LLMs. Empirical evaluations demonstrate that the LLaMA-3 model, with an MAE of 0.216 and an NDCG@20 of 0.901, significantly surpasses the performance of prior methods when solely relying on internal information. Furthermore, the impact values predicted by our method show a strong positive correlation with the quartile rankings of journals for articles published in 2024, illustrating the practical applicability of our approach in real-world settings. Overall, the proposed approach effectively estimates a future impact score from 0 to 1 for newly published papers, presenting considerable benefits for individuals, institutions, and automated scientific research systems.

\appendix
\section*{Appendices}

\subsection{Implementation Details}
% % All the code for this study are written with Python. We adopt MySQL 8.0 for constructing the datasets, and train the language models with libraries including PyTorch, Transformers, and PEFT, \textit{etc}. Detailed version information can be found in our open-source repository.

In this paper, We partition the NAID into training, validation, and test sets in a ratio of 8:1:1. We conduct a grid search on the validation set and identify optimal settings for several key hyperparameters. The initial learning rate is set to 5e-5 and is dynamically adjusted based on the effective batch size and training steps. The maximum length for the LLM is 1024. The rank $r$ for LoRA is set to 16 and is only applied to the $q$ and $v$ matrices in the self-attention mechanism. To further reduce memory consumption, we employ 8-bit model quantization techniques. All experiments are carried out on the server with 8 × NVIDIA A40 GPUs, with fine-tuning completed in less than two hours (or roughly 20 hours on a single RTX 3090 GPU). Unless otherwise stated, all models are trained for 5 epochs. All the metrics mentioned in this paper are calculated in July 2024. Please note that the citation counts provided by Semantic Scholar are typically lower than those displayed by Google Scholar. More experiment settings could be found in the provided code framework.

\subsection{Will Additional Information Boost Performance?}
We have already demonstrated that LLMs are capable of predicting future impact using only titles and abstracts. This raises the question of whether additional information not included in the abstract could further enhance prediction performance. Therefore, we design multiple experiments to quantitatively analyze whether the availability of a paper's open-source code, achievement of SOTA performance, contribution of a new dataset, and the quality of its references~\cite{zhao2024literature} may boost the performance. The additional information is extracted by gpt-3.5-turbo-0125, which reads as much of the article text as possible.
\begin{table}[ht]
    
    \begin{center}
        \begin{tabular}{p{0.5\linewidth}p{0.15\linewidth}p{0.15\linewidth}}
            \hline
            Input Type &  MAE↓ & NDCG↑  \\
            \hline
            \hline
             Title\&Abstract & 0.216 & 0.901  \\
            + Open Access Status  & 0.214 & 0.878  \\
            + Release Dataset  & 0.212 & 0.915  \\
            + SOTA Claim  & 0.215 & 0.881  \\
            + RQM~\cite{zhao2024literature}   & 0.211 & \textbf{0.931}  \\
            + All above  & \textbf{0.209} & 0.917  \\
            \hline

        \end{tabular}
    \end{center}
    \caption{Ablation study on additional information: additional information boosts prediction performance.}
    \label{tab:additional_info}
\end{table}

The prompt templated is slightly modified from the original one: ``\textit{Given a certain paper, Title: \{title\} Abstract: \{abstract\}. State-of-the-Art Performance: \{'Yes' or 'No'\}.  Released a New Dataset: \{'Yes' or 'No'\}.  Code Open Access: \{'Yes' or 'No'\}. Reference Quality Metric(on a scale from lowest 0 to highest 1): \{RQM\}. Predict its normalized academic impact (between 0 and 1):}''.

As observed in Tab.~\ref{tab:additional_info}, nearly all additional information contributes to an improvement in the MAE metric. However, only the availability of an open-source dataset and the quality of references positively impact the NDCG metric. This effect may be attributed to the fact that abstracts typically already include key information, such as SOTA performance, and adding redundant data may bring unnecessary complexity to the LLM. Nevertheless, when the model is provided with all of the information, both the MAE and NDCG metrics exhibit improvements compared to scenarios where no additional information is included. This suggests that the incorporation of additional information may enhance overall performance.

\subsection{Impact of Different Training Schemes on Predictive Performance}
In our paper, we adopt one of the most common methods for fine-tuning LLMs, LoRA~\cite{hu2021lora}. This section explores how other fine-tuning approaches may influence performance. As shown in Tab.~\ref{tab:FT_type}, we also experiment with several fine-tuning methods, including freezing the backbone and fine-tuning only the classification head MLP, as well as employing rsLoRA~\cite{kalajdzievski2023rank}, and DoRA~\cite{liu2024dora} to fine-tune the model.

\begin{table}[ht]
    
    \begin{center}
        \begin{tabular}{p{0.5\linewidth}p{0.15\linewidth}p{0.15\linewidth}}
            \hline
            Training schemes &  MAE↓ & NDCG↑  \\
            \hline
            \hline
            LoRA~\cite{hu2021lora} & 0.216 & 0.901  \\
            
            Classification Head Only & 0.237 & 0.839  \\
            
            rsLoRA~\cite{kalajdzievski2023rank}   & \textbf{0.213} & 0.897  \\
            
            DoRA~\cite{liu2024dora}  & 0.217 & \textbf{0.902}  \\
            \hline

        \end{tabular}
    \end{center}
    \caption{Comparison of Various Training Schemes: fine-tuning with LoRA yields better results than fine-tuning only the classification head.} 
    \label{tab:FT_type}
\end{table}

\subsection{Impact of the Various Loss Function on Predictive Performance}

We investigate the impact of different loss functions on model performance. In addition to MSE loss, we test the performance when using L1, SmoothL1, and BCELoss as the loss functions. 
It is crucial to note that for BCELoss, we employ PyTorch's built-in BCEWithLogitsLoss during training to enhance numerical stability. The sigmoid function is applied separately during the inference phase to normalize the outputs.

The experimental results are presented in Tab.~\ref{tab:loss_compare}, showing that the model’s ability to identify high-impact papers is strongest when MSE is used as the loss function. The performance of L1 and SmoothL1 is similar, which may be due to the balanced nature of the NAID dataset. BCELoss performs slightly worse than SmoothL1, with an NDCG score of 0.762.

\begin{table}[ht]
    
    \begin{center}
        \begin{tabular}{p{0.4\linewidth}p{0.15\linewidth}}
            \hline
            Loss Function  & NDCG↑  \\
            \hline
            \hline
            MSE Loss & \textbf{0.901}  \\
            
            L1 Loss  & 0.831  \\
            
            SmoothL1 Loss& 0.787  \\
            
            BCE Loss & 0.762  \\
            \hline

        \end{tabular}
    \end{center}
    \caption{Comparison of adopting various loss functions: MSE loss delivers the best performance in terms of NDCG.}
    \label{tab:loss_compare}
\end{table}

% \section{Predicted Impact of this paper}
% Another fascinating application of the proposed method is rephrasing titles and abstracts based on the predicted TNCSI$\mathrm{_\text{SP}}$. Such approach poses fewer ethical risks compared to directly using LLMs (\textit{e.g.,} ChatGPT) for refining titles and abstracts. As shown in Tab. X, We test various versions of titles and abstracts and determine that the current one is the most likely to capture readers' interest and potentially inspire more outstanding work in the future.

\subsection{Reproduction Details of Previous Methods}
To demonstrate the superiority of our proposed method, we compare it with previous SOTA methods in the Approach section. In this subsection, we further explain how we replicate these methods.

As previously mentioned, article impact prediction methods at different levels rely on varying data and exhibit different task complexities. Directly comparing methods across all levels may lead to potential unfairness. Therefore, we excluded external features such as journals characteristics, citation features, as well as author and affiliation reputation when making predictions for level I methods. 

To mitigate potential issues such as gradient explosion, we normalize the inputs for all MLP-based methods to ensure that different features are on a consistent scale. Specifically, for the work by Hu et al.~\cite{hu2023identifying}, which primarily focuses on classifying whether an article belongs to the notable top 5\%, we convert their approach into a regression task that represents cumulative impact (TNCSI). For the approach proposed by Zhang et al.~\cite{zhang2024predicting}, we divide the training data into three groups based on arXiv category IDs: cs.CV, cs.CL, and cs.AI. Then, we train separate models for each group and utilize the corresponding model weights for inference based on the category of the data during testing. For the ChatGPT-based method~\cite{de2024can}, we replicated the approach using the exact prompts provided by the authors. Surprisingly, the original performance of ChatGPT in predicting article impact is unsatisfactory (about an NDCG of 0.05). This may be due to the fact that the original method is not specifically designed for impact prediction but rather aims to investigate the correlation between ChatGPT's scores and various factors, and is not specifically designed to predict scores that reflect academic impact. Therefore, we emulate the prompt used by LLaMA3 to guide ChatGPT (gpt-3.5-turbo-0125) in generating responses, which significantly increases the corresponding NDCG value to 0.597. LLaMA-3-generated refers to employing LLaMA-3 in a chat-based manner, where the pre-trained LLaMA-3 model generates autoregressive outputs to predict influence.

% , as evidenced by the claimed weak correlation between ChatGPT's scores and citation counts. Nevertheless, we find this approach intriguing and believe it has the potential to inspire further research into zero-shot article impact prediction.

Due to the specialized nature of the field, most papers are designed to serve specific institutions, resulting in limited availability of open-source code. Additionally, the high cost of accessing certain databases has further hindered our ability to reproduce experiments from some papers. Consequently, despite our best efforts to replicate the reported methods, there may still be minor differences in the details compared to the original authors' approaches. Nevertheless, we maintain that these discrepancies are unlikely to have a significant impact on the overall experimental outcomes.

\subsection{Preview of the TKPD and NAID}
We provide a preview of the TKPD and NAID datasets in Tab.~\ref{tab:TKPD_preview} and~\ref{tab:NAID_preview}, respectively. The TKPD dataset includes over 200 entries, consisting of titles, abstracts, and manually annotated topic key phrases. The NAID dataset comprises over 12,000 entries, featuring fields such as titles, abstracts, $\mathrm{\text{TNCSI}}$, TNCSI$_\mathrm{\text{SP}}$, open-source status, and RQM values, \textit{etc}. Both datasets are available in our Supplementary Material and project page.

\begin{table}[ht]
    
    \begin{center}
        \begin{tabular}{p{0.6\linewidth}p{0.25\linewidth}}
            \hline
            Title &  Topic  \\
            \hline
            \hline
             Oracle-MNIST: a Dataset of Oracle Characters for Benchmarking Machine Learning Algorithms & oracle character recognition  \\
            \hline
            Bridging Cross-Lingual Gaps During Leveraging the Multilingual Sequence-to-Sequence Pretraining for Text Generation and Understanding  & cross-lingual text generation \\
            \hline
            UniSAr: A Unified Structure-Aware Autoregressive Language Model for Text-to-SQL & text-to-SQL  \\
            \hline
            Inpainting at Modern Camera Resolution by Guided PatchMatch with Auto-Curation & image inpainting  \\
            \hline
        \end{tabular}
    \end{center}
    \caption{A preview of the TKPD: for clarity of presentation, we have excluded the abstract field.}
    \label{tab:TKPD_preview}
\end{table}

\begin{table*}[ht]
    
    \begin{center}
        \begin{tabular}{p{0.3\linewidth}p{0.42\linewidth}p{0.05\linewidth}p{0.08\linewidth}}
            \hline
            Title & Abstract & Cites &  $\mathrm{\text{TNCSI}}_\mathrm{\text{SP}}$ \\
            \hline
            \hline
%             Training language models to follow instructions with human feedback & Making language models bigger does not inherently make them better at
% following a user's intent. For example,...& 6131  & 1\\
%             \hline
            LoRA: Low-Rank Adaptation of Large Language Models & An important paradigm of natural language processing consists of  ...  & 4421 & 1\\
            \hline
            HuBERT: Self-Supervised Speech Representation Learning by Masked Prediction of Hidden Units & Self-supervised approaches for speech representation learning are challenged by three unique problems: (1) there are multiple ... & 1793 & 1 \\
            \hline
            YOLOX: Exceeding YOLO Series in 2021 & In this report, we present some experienced improvements to YOLO series ... & 2651 & 0.906 \\
            \hline
            MobileBERT: a Compact Task-Agnostic BERT for Resource-Limited Devices & Natural Language Processing (NLP) has recently achieved great success by using huge pre-trained models with ... & 665 & 0.963 \\
            \hline
            A Time Series is Worth 64 Words: Long-term Forecasting with Transformers & We propose an efficient design of Transformer-based models for multivariate
time series forecasting and self-supervised representation learning. It ... & 298 & 0.825 \\
            \hline 
            XLM-T: Multilingual Language Models in Twitter for Sentiment Analysis and Beyond & Language models are ubiquitous in current NLP, and their multilingual capacity has recently attracted considerable attention. However, ...  & 152 & 0.426 \\
            % \hline 
            % OpenScene: 3D Scene Understanding with Open Vocabularies & Traditional 3D scene understanding approaches rely on labeled 3D datasets to train a model for a single task with supervision. We propose OpenScene, ... & 149 & 0.847 \\
            \hline
            Confidence Score for Source-Free Unsupervised Domain Adaptation & Source-free unsupervised domain adaptation (SFUDA) aims to obtain high performance ... & 42 & 0.299 \\
            \hline
            An Improved Dilated Convolutional Network for Herd Counting in Crowded Scenes & Crowd management technologies that leverage computer vision are widespread in
contemporary times. There exists many ...  & 2 & 0.004 \\

        \hline
        \end{tabular}
    \end{center}
    \caption{A preview of the NAID: compared to citation counts, the proposed $\mathrm{\text{TNCSI}}_\mathrm{\text{SP}}$ metric offers cross-domain comparability. For instance, although YOLOX in the field of object detection has received an impressive 2,600 citations, its $\mathrm{\text{TNCSI}}_\mathrm{\text{SP}}$ is slightly lower than that of MobileBERT. Such a phenomenon could be attributed to the larger researcher base and higher average citations in object detection at that time. Only key fields are shown for clarity.}
    \label{tab:NAID_preview}
\end{table*}

\section*{Ethical Statement}
We are aware of the potential for manipulation through excessive optimization of titles and abstracts. Researchers must refrain from excessively embellishing titles and abstracts, particularly by making false claims about unachieved performance or overly exaggerating the significance of their methods, in an attempt to manipulate predicted impact values. 

Due to constraints such as the access frequency limits of the Semantic Scholar API, we are unable to construct a larger dataset. Therefore, our proposed method only serves as a preliminary exploratory approach. The predictions generated by this method are probabilistic estimates and should never be considered definitive assessments of an article's quality. The method is intended to provide additional insights and must not replace the existing peer-review process, which remains essential for maintaining the integrity and rigor of academic research. The authors are not responsible for any decisions made based on the predictions.

\section*{Acknowledgements}
This research was supported by the Young Scientists Fund of the National Natural Science Foundation of China (Grant No.62206134), the Fundamental Research Funds for the Central Universities 070-63233084, and the Tianjin Key Laboratory of Visual Computing and Intelligent Perception (VCIP). Computation is supported by the Supercomputing Center of Nankai University (NKSC). This work was supported by the National Science Fund of China under Grant No. 62361166670.

\begin{quote}
\begin{small}

\bibliography{references}
\end{small}
\end{quote}

\section{Reproducibility Checklist}
The code and dataset (detailed in the Approach section) are released in the Supplementary Material and on our project webpage. As requested by \url{https://aaai.org/conference/aaai/aaai-25/aaai-25-reproducibility-checklist/}, this paper:
\begin{itemize}

\item Includes a conceptual outline and/or pseudocode description of AI methods introduced. \textbf{Yes.}

\item Clearly delineates statements that are opinions, hypothesis, and speculation from objective facts and results. \textbf{Yes.}

\item Provides well marked pedagogical references for less-familiare readers to gain background necessary to replicate the paper. \textbf{Yes.}
\end{itemize}

\noindent Does this paper make theoretical contributions? \textbf{No.}

\noindent Does this paper rely on one or more datasets? \textbf{Yes.}
\begin{itemize}
\item A motivation is given for why the experiments are conducted on the selected datasets. \textbf{Yes.}
\item All novel datasets introduced in this paper are included in the Supplementary Materials. \textbf{Yes.}
\item All novel datasets introduced in this paper will be made publicly available upon publication of the paper with a license that allows free usage for research purposes. \textbf{Yes.}
\item All datasets drawn from the existing literature (potentially including authors’ own previously published work) are accompanied by appropriate citations. \textbf{NA.}
\item All datasets drawn from the existing literature (potentially including authors’ own previously published work) are publicly available. \textbf{NA.}
\item All datasets that are not publicly available are described in detail, with explanation why publicly available alternatives are not scientifically satisficing. \textbf{NA.}
\end{itemize}
\noindent Does this paper include computational experiments? \textbf{Yes.}
\begin{itemize}
\item Any code required for pre-processing data is included in the Supplementary Materials. \textbf{Yes.}
\item All source code required for conducting and analyzing the experiments is included in a Supplementary Materials. \textbf{Yes.}
\item All source code required for conducting and analyzing the experiments will be made publicly available upon publication of the paper with a license that allows free usage for research purposes. \textbf{Yes.}
\item All source code implementing new methods have comments detailing the implementation, with references to the paper where each step comes from. \textbf{Yes.}
\item If an algorithm depends on randomness, then the method used for setting seeds is described in a way sufficient to allow replication of results.  \textbf{NA.}
\item This paper specifies the computing infrastructure used for running experiments (hardware and software), including GPU/CPU models; amount of memory; operating system; names and versions of relevant software libraries and frameworks. \textbf{Yes.}
\item This paper formally describes evaluation metrics used and explains the motivation for choosing these metrics. \textbf{Yes.}
\item This paper states the number of algorithm runs used to compute each reported result. \textbf{Yes.}
\item Analysis of experiments goes beyond single-dimensional summaries of performance (e.g., average; median) to include measures of variation, confidence, or other distributional information. \textbf{No.}
\item The significance of any improvement or decrease in performance is judged using appropriate statistical tests (e.g., Wilcoxon signed-rank).\textbf{No.}
\item This paper lists all final (hyper-)parameters used for each model/algorithm in the paper’s experiments. \textbf{Partial.}
\item This paper states the number and range of values tried per (hyper-) parameter during development of the paper, along with the criterion used for selecting the final parameter setting. \textbf{Partial.}
\end{itemize}
\clearpage

\end{document}